\documentclass{article}
\usepackage[dvips]{graphicx}
\usepackage{arxiv}
\usepackage[utf8]{inputenc} 
\usepackage[T1]{fontenc}    
\usepackage{hyperref}       
\usepackage{breakurl}       
\usepackage{booktabs}       
\usepackage{amsfonts}       
\usepackage{nicefrac}       
\usepackage{microtype}      
\usepackage{bm}
\usepackage{lipsum}
\usepackage{amsmath}
\usepackage{algorithm}
\usepackage{algorithmic}
\usepackage{multirow}
\usepackage{comment}
\title{An On-Device Federated Learning Approach for Cooperative
  Model Update between Edge Devices}
\author{
  Rei Ito\\
  Keio University\\
  3-14-1 Hiyoshi, Kohoku-ku, Yokohama, Japan\\
  \texttt{rei@arc.ics.keio.ac.jp}\\
  \And
  Mineto Tsukada\\
  Keio University\\
  3-14-1 Hiyoshi, Kohoku-ku, Yokohama, Japan\\
  \texttt{tsukada@arc.ics.keio.ac.jp}\\
  \And
  Hiroki Matsutani \\
  Keio University\\
  3-14-1 Hiyoshi, Kohoku-ku, Yokohama, Japan\\
  \texttt{matutani@arc.ics.keio.ac.jp} \\
}

\begin{document}
\maketitle

\begin{abstract}
Most edge AI focuses on prediction tasks on resource-limited edge
devices while the training is done at server machines.
However, retraining or customizing a model is required at
edge devices as the model is becoming outdated due to environmental
changes over time.
To follow such a concept drift, a neural-network based on-device
learning approach is recently proposed, so that edge devices train
incoming data at runtime to update their model.
In this case, since a training is done at distributed edge devices,
the issue is that only a limited amount of training data can be used
for each edge device.
To address this issue, one approach is a cooperative learning or
federated learning, where edge devices exchange their trained results
and update their model by using those collected from the other devices.
In this paper, as an on-device learning algorithm, we focus on OS-ELM
(Online Sequential Extreme Learning Machine) to
sequentially train a model based on recent samples and
combine it with autoencoder for anomaly detection.
We extend it for an on-device federated learning so that edge devices
can exchange their trained results and update their model by using those
collected from the other edge devices.
This cooperative model update is one-shot while it can be
repeatedly applied to synchronize their model.
Our approach is evaluated with anomaly detection tasks
generated from a driving dataset of cars, a human activity dataset,
and MNIST dataset.
The results demonstrate that the proposed on-device federated learning
can produce a merged model by integrating trained results from
multiple edge devices
as accurately as traditional backpropagation based neural
networks and a traditional federated learning approach with lower
computation or communication cost.
\end{abstract}


\section{Introduction}\label{sec:intro}

%
%
Most edge AI focuses on prediction tasks on resource-limited edge
devices assuming that their prediction model has been trained at
server machines beforehand.
 However, retraining or customizing a model is
required at edge devices as the model is becoming outdated due to
environmental changes over time (i.e., concept drift). 
Generally, retraining the model later to reflect environmental changes
for each edge device is a complicated task, because the server machine
needs to collect training data from the edge device, train a new model
based on the collected data, and then deliver the new model to the
edge device.

%
%
 To enable the retraining a model at resource-limited edge
devices, in this paper we use a neural network based on-device
learning approach \cite{Tsukada18,Tsukada20} since it can sequentially train
neural networks at resource-limited edge devices and also the neural
networks typically have a high flexibility to address various
nonlinear problems. 
Its low-cost hardware implementation is also introduced in
\cite{Tsukada20}.
In this case, since a training is done independently at distributed
edge devices, the issue is that only a limited amount of training data
can be used for each edge device.
To address this issue, one approach is a cooperative model update,
where edge devices exchange their trained results and update their
model using those collected from the other devices.
Here we assume that edge devices share an intermediate form of their
weight parameters instead of raw data, which is sometimes privacy
sensitive.

%
%
 In this paper, we use the on-device learning approach
\cite{Tsukada18,Tsukada20} based on OS-ELM (Online Sequential Extreme
Learning Machine) \cite{Liang06} and autoencoder \cite{Hinton06}.
Autoencoder is a type of neural network architecture which can be
applied to unsupervised or semi-supervised anomaly detection, and
OS-ELM is used to sequentially train neural networks at resource-limited edge
devices. 
It is then extended for the on-device federated learning so that edge
devices can exchange their trained results and update their model
using those collected from the other edge devices.
 In this paper, we
employ a concept of Elastic ELM (E$^2$LM) \cite{Xin15}, which is a
distributed training algorithm for ELM (Extreme Learning Machine)
\cite{Huang04}, so that intermediate training results are computed by
edge devices separately and then a final model is produced by
combining these intermediate results. 
It is applied to the OS-ELM based on-device learning approach to
construct the on-device federated learning.
 Please note that although in this paper the on-device
federated learning is applied to anomaly detection tasks since the
baseline on-device learning approach \cite{Tsukada18,Tsukada20} is
designed for anomaly detection tasks, the proposed approach that
employs the concept of E$^2$LM is more general and can be applied to
the other machine learning tasks. 
In the evaluations, we will demonstrate that the proposed on-device
federated learning can produce a merged model by integrating trained
results from multiple edge devices 
 as accurately as traditional backpropagation based neural
networks and a traditional federated learning approach with lower
computation or communication cost
\footnote{This preprint has been accepted for IEEE Access,
DOI:10.1109/ACCESS.2021.3093382 \cite{Ito21}.}.

%
%
The rest of this paper is organized as follows.
Section \ref{sec:relate} overviews traditional federated learning
technologies.
Section \ref{sec:prelim} introduces baseline technologies behind the
proposed on-device federated learning approach.
Section \ref{sec:design} proposes a model exchange and update
algorithm of the on-device federated learning.
Section \ref{sec:eval} evaluates the proposed approach using 
three datasets in terms of accuracy and latency.
Section \ref{sec:conc} concludes this paper.

\section{Related Work}\label{sec:relate}
%
%
A federated learning framework was proposed by Google in 2016
\cite{Konecny16,Konecny16-v2,Brendan16}.
 Their main idea is to build a global federated model at a
server side by collecting locally trained results from distributed
client devices. 
In \cite{Brendan16}, a secure client-server structure that can avoid
information leakage is proposed for federated learning. 
More specifically, Android phone users train their models locally 
and then the model parameters are uploaded to the server side in a
secure manner.

%
%
Preserving of data privacy is an essential property for federated
learning systems.
 In \cite{Shokri15}, a collaborative deep learning scheme
where participants selectively share their models' key parameters is
proposed in order to keep their privacy. 
In the federated learning system, participants compute gradients
independently and then upload their trained results to a parameter
server.
As another research direction, information leakage at the server side 
is discussed by considering data privacy and security issues. 
Actually, a leakage of these gradients may leak important data 
when the data structure or training algorithm is exposed simultaneously. 
To address this issue, in \cite{Phong18}, an additively homomorphic
encryption is used for masking the gradients in order to preserve
participants' privacy and enhance the security at the server side.

%
%
Recently, some prior work involved in federated learning focuses on
the communication cost or performance in massive or unbalanced data
distribution environments.   
In \cite{Lin17}, a compression technique called Deep Gradient
Compression is proposed for large-scale distributed training in order
to reduce the communication bandwidth.

%
%
A performance of centralized model built by a federated learning
system depends on statistical nature of data collected from 
client devices.
Typically, data in the client side is not always independent and
identically distributed (IID), because clients' interest and
environment are different and sometimes degrade the model performance.
In \cite{Zhao18}, it is shown that accuracy of a federated learning 
is degraded for highly skewed Non-IID data. 
This issue is addressed by creating a small subset of data which is
globally shared between all the clients.
 In \cite{Shoham19}, it is reported that locally trained
models may be forgot by a federated learning with Non-IID data, and a
penalty term is added to a loss function to prevent the knowledge
forgetting. 

%
%
As a common manner, a server side in federated learning systems
has no access to local data in client devices.  
There is a risk that a client may get out of normal behaviors in the
federated model training. 
In \cite{Li19}, a dimensionality reduction based anomaly detection
approach is utilized to detect anomalous model updates from
clients in a federated learning system.
 In \cite{Shen16}, malicious clients are identified by
clustering their submitted features, and then the final global model
is generated by excluding updates from the malicious clients. 

%
%
 Many existing federated learning systems assume
backpropagation based sophisticated neural networks but their training
is compute-intensive.
In our federated learning approach, although we also use neural
networks, we employ a recently proposed on-device learning approach
for resource-limited edge devices, which will be introduced in the
next section. 
Also, please note that in our approach we assume that intermediate
training results are exchanged via a server for simplicity; however,
local training and merging from intermediate training results from
other edge devices can be completed at each edge device.


\section{Preliminaries}\label{sec:prelim}

\begin{figure}[t]
    \centering
    \includegraphics[height=40mm]{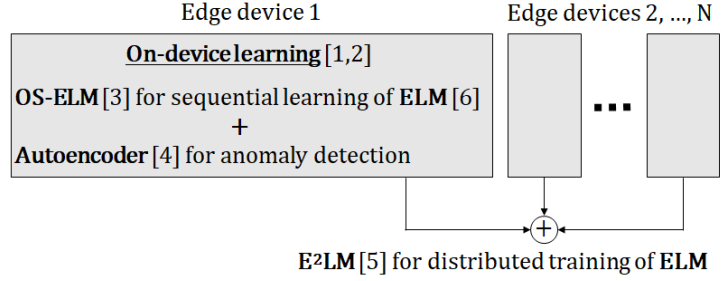}
    \caption{Baseline technologies behind our proposal}
    \label{fig:overview}
\end{figure}

This section briefly introduces baseline technologies behind our
proposal: 1) ELM (Extreme Learning Machine), 2) E$^2$LM (Elastic
Extreme Learning Machine), 3) OS-ELM (Online Sequential Extreme
Learning Machine), and 4) autoencoder.
Figure \ref{fig:overview} illustrates the proposed
cooperative model update between $N$ edge devices, each of which
performs the on-device learning that combines OS-ELM and autoencoder.
Their intermediate training results are merged by using E$^2$LM. 
Note the original E$^2$LM algorithm is designed for ELM, not OS-ELM; 
so we modified it so that trained results of OS-ELM are merged, which
will be shown in Section \ref{sec:design}.

\subsection{ELM}\label{ssec:elm}

\begin{figure}[t]
    \centering
    \includegraphics[height=60mm]{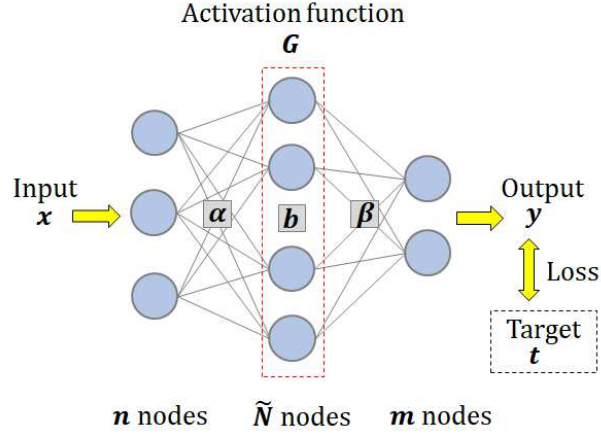}
    \caption{Single hidden-layer feedforward network (SLFN)}
    \label{fig:elm}
\end{figure}

%
%
ELM \cite{Huang04} is a batch training algorithm for single
hidden-layer feedforward networks (SLFNs).
As shown in Figure \ref{fig:elm}, the network consists of input layer,
hidden layer, and output layer.
The numbers of their nodes are denoted as $n$, $\tilde{N}$, and $m$,
respectively. 
Assuming an $n$-dimensional input chunk 
$\bm{x} \in \bm{R}^{k \times n}$ of batch size $k$ is given, 
an $m$-dimensional output chunk $\bm{y} \in \bm{R}^{k \times m}$ is
computed as follows.
\begin{equation}\label{eq:elm_predict}
    \bm{y} = G(\bm{x} \cdot \bm{\alpha} + \bm{b})\bm{\beta},
\end{equation}
where $G$ is an activation function, 
$\bm{\alpha}\in\bm{R^{n \times \tilde{N}}}$ is an input weight matrix
between the input and hidden layers, 
$\bm{\beta}\in\bm{R^{\tilde{N} \times m}}$ is an output weight matrix
between the hidden and output layers, and
$\bm{b}\in\bm{R^{\tilde{N}}}$ is a bias vector of the hidden layer. 

%
%
If an SLFN model can approximate $m$-dimensional target chunk (i.e.,
teacher data) $\bm{t} \in \bm{R}^{k \times m}$ with zero error
($Loss=0$), the following equation is satisfied.
\begin{equation}\label{eq:elm_zero_error}
    G(\bm{x} \cdot \bm{\alpha} + \bm{b})\bm{\beta} = \bm{t}
\end{equation}
Here, the hidden-layer matrix is defined as 
$\bm{H} \equiv G(\bm{x} \cdot \bm{\alpha} + \bm{b})$.
The optimal output weight matrix $\hat{\bm{\beta}}$ is computed as follows.
\begin{equation}\label{eq:elm_train}
    \hat{\bm{\beta}} = \bm{H}^{\dagger}\bm{t},
\end{equation}
where $\bm{H}^{\dagger}$ is a pseudo inverse matrix of $\bm{H}$, 
which can be computed with matrix decomposition algorithms, such as
SVD (Singular Value Decomposition) and QRD (QR Decomposition).

%
%
In ELM algorithm, the input weight matrix $\bm{\alpha}$ is initialized
with random values and not changed thereafter.
The optimization is thus performed only for the output weight matrix 
$\bm{\beta}$, and so it can reduce the computation cost compared with
backpropagation based neural networks that optimize both $\bm{\alpha}$
and $\bm{\beta}$.
In addition, the training algorithm of ELM is not iterative; it
analytically computes the optimal weight matrix $\bm{\beta}$ for a given
input chunk in a one-shot manner, as shown in Equation
\ref{eq:elm_train}. 
It can always obtain a global optimal solution for $\bm{\beta}$,
unlike a typical gradient descent method, which sometimes converges to
a local optimal solution.

%
%
Please note that ELM is one of batch training algorithms for SLFNs,
which means that the model is trained by using all the training data
for each update.
In other words, we need to retrain the whole data in order to update
the model for newly-arrived training samples.
This issue is addressed by E$^2$LM and OS-ELM.

\subsection{E$^2$LM}\label{ssec:e2lm}

%
%
E$^2$LM \cite{Xin15} is an extended algorithm of ELM for enabling the
distributed training of SLFNs.
That is, intermediate training results are computed by multiple
machines separately, and then a merged model is produced by combining
these intermediate results.

%
%
In Equation \ref{eq:elm_train}, assuming that rank $\bm{H} = \tilde{N}$ 
and $\bm{H^T}\bm{H}$ is nonsingular, the pseudo inverse matrix 
$\bm{H}^{\dagger}$ is decomposed as follows.
\begin{equation}\label{equation_decomp}
    \bm{H^{\dagger}} = {(\bm{H^T}\bm{H})}^{-1}\bm{H^T}
\end{equation}
The optimal output weight matrix $\bm{\beta}$ in Equation
\ref{eq:elm_train} can be computed as follows.
\begin{equation}\label{equation_pinv2}
    \hat{\bm{\beta}} = {(\bm{H^T}\bm{H})}^{-1}\bm{H^T}\bm{t}
\end{equation}
Assuming the intermediate results are defined as
$\bm{U} = \bm{H^T}\bm{H}$ and $\bm{V} = \bm{H^T}\bm{t}$, the above
equation is denoted as follows.
\begin{equation}\label{equation_pinv3}
    \hat{\bm{\beta}} = \bm{U}^{-1}\bm{V}
\end{equation}

%
%
Here, the hidden-layer matrix and target chunk (i.e., teacher data)
for newly-arrived training dataset $\Delta\bm{x}$ are denoted as
$\Delta\bm{H}$ and $\Delta\bm{t}$, respectively.
The intermediate results for $\Delta\bm{x}$ are denoted as
$\Delta\bm{U} = \Delta\bm{H^T}\Delta\bm{H}$ and 
$\Delta\bm{V} = \Delta\bm{H^T}\Delta\bm{t}$. 

%
%
Similarly, the hidden-layer matrix and target chunk for updated
training dataset $\bm{x}^{\prime} = \bm{x} + \Delta\bm{x}$ are denoted
as $\bm{H}^{\prime}$ and $\bm{t}^{\prime}$, respectively.
The intermediate results for $\bm{x}^{\prime}$ are denoted as
$\bm{U}^{\prime} = \bm{H}^{\prime \bm{T}}\bm{H^{\prime}}$ and
$\bm{V^{\prime}} = \bm{H}^{\prime \bm{T}}\bm{t^{\prime}}$. 
Then, $\bm{U}^{\prime}$ and $\bm{V}^{\prime}$ can be computed as
follows.
\begin{equation}\label{equation_uv}
    \begin{split}
        \bm{U}^{\prime} &= \bm{H}^{\prime \bm{T}}\bm{H^{\prime}} = {\begin{bmatrix} \bm{H} \\ \Delta\bm{H}\end{bmatrix}}^{T}\begin{bmatrix} \bm{H} \\ \Delta\bm{H}\end{bmatrix} = \bm{H^T}\bm{H} + \Delta\bm{H^T}\Delta\bm{H}\\
        \bm{V}^{\prime} &= \bm{H}^{\prime \bm{T}}\bm{t^{\prime}} = {\begin{bmatrix} \bm{H} \\ \Delta\bm{H}\end{bmatrix}}^{T}\begin{bmatrix} \bm{t} \\ \Delta\bm{t}\end{bmatrix} = \bm{H^T}\bm{t} + \Delta\bm{H^T}\Delta\bm{t}
    \end{split}
\end{equation}
As a result, Equation \ref{equation_uv} can be denoted as follows.
\begin{equation}\label{equation_uv2}
    \begin{split}
        \bm{U}^{\prime} = \bm{U} + \Delta\bm{U}\\
        \bm{V}^{\prime} = \bm{V} + \Delta\bm{V}
    \end{split}
\end{equation}

%
%
In summary, E$^2$LM algorithm updates a model in the following steps:
\begin{enumerate}
\item Compute $\bm{U}$ and $\bm{V}$ for the whole training dataset $\bm{x}$,
\item Compute $\Delta\bm{U}$ and $\Delta\bm{V}$ for newly-arrived
  training dataset $\Delta\bm{x}$, 
\item Compute $\bm{U}^{\prime}$ and $\bm{V}^{\prime}$ for updated
  training dataset $\bm{x}^{\prime}$ using Equation \ref{equation_uv2}, and
\item Compute the new output weight matrix $\bm{\beta}$ using Equation
  \ref{equation_pinv3}.
\end{enumerate}
Please note that we can compute a pair of $\bm{U}$ and $\bm{V}$ and a
pair of $\Delta\bm{U}$ and $\Delta\bm{V}$ separately.
Then, we can produce $\bm{U}^{\prime}$ and $\bm{V}^{\prime}$ by simply
adding them using Equation \ref{equation_uv2}.
Similar to the addition of $\bm{x}$ and $\Delta\bm{x}$, subtraction
and replacement operations for $\bm{x}$ are also supported.

\subsection{OS-ELM}\label{ssec:oselm}

OS-ELM \cite{Liang06} is an online sequential version of ELM, which
can update the model sequentially using an arbitrary batch size.

Assuming that the $i$-th training chunk 
$\{\bm{x}_i \in \bm{R}^{k_i \times n}, \bm{t}_i \in \bm{R}^{k_i \times m}\}$
of batch size $k_i$ is given, we need to compute the output weight
matrix $\bm{\beta}$ that can minimize the following error.
\begin{equation}\label{eq:os_elm_error}
    \begin{Vmatrix}
        \begin{bmatrix}\bm{H}_0 \\ \vdots \\ \bm{H}_i\end{bmatrix} \bm{\beta}_i - \begin{bmatrix} \bm{t}_0 \\ \vdots \\ \bm{t}_i \end{bmatrix}
    \end{Vmatrix},
\end{equation}
where $\bm{H}_i$ is defined as 
$\bm{H}_i \equiv G(\bm{x}_i \cdot \bm{\alpha} + \bm{b})$.
Assuming 
\begin{equation}
\bm{K_i} \equiv {\begin{bmatrix} \bm{H_0} \\ \vdots \\ \bm{H_i}\end{bmatrix}}^{T}\begin{bmatrix} \bm{H_0} \\ \vdots \\ \bm{H_i}\end{bmatrix} (i \ge 0), 
\end{equation}
the optimal output weight matrix is computed as follows.
\begin{equation}\label{eq:os_elm_train1}
    \begin{split}
        \bm{\beta_{i}} &= \bm{\beta_{i-1}} + \bm{K_{i}^{-1}}\bm{H_{i}^T}(\bm{t_{i}} - \bm{H_{i}}\bm{\beta_{i-1}})\\
        \bm{K_{i}} &= \bm{K_{i-1}} + \bm{H_{i}^T}\bm{H_{i}}
    \end{split}
\end{equation}
Assuming $\bm{P_i} \equiv \bm{K_i^{-1}}$, we can derive the following
equation from Equation \ref{eq:os_elm_train1}.
\begin{equation}\label{eq:os_elm_train2}
    \begin{split}
        \bm{P}_i &= \bm{P}_{i-1} - \bm{P}_{i-1}\bm{H}_i^T(\bm{I} + \bm{H}_i\bm{P}_{i-1}\bm{H}_i^T)^{-1}\bm{H}_i\bm{P}_{i-1}\\
        \bm{\beta}_i &= \bm{\beta}_{i-1} + \bm{P}_i\bm{H}_i^T(\bm{t}_i - \bm{H}_i\bm{\beta}_{i-1})
    \end{split}
\end{equation}
In particular, the initial values $\bm{P}_0$ and $\bm{\beta}_0$ are
precomputed as follows.
\begin{equation}\label{eq:os_elm_init}
    \begin{split}
        \bm{P}_0 &= (\bm{H}_0^T\bm{H}_0)^{-1}\\
        \bm{\beta}_0 &= \bm{P}_0\bm{H}_0^T\bm{t}_0
    \end{split}
\end{equation}
As shown in Equation \ref{eq:os_elm_train2}, the output weight matrix 
$\bm{\beta}_i$ and its intermediate result $\bm{P}_i$ are computed
from the previous training results $\bm{\beta}_{i-1}$ and $\bm{P}_{i-1}$. 
Thus, OS-ELM can sequentially update the model with a newly-arrived
target chunk in a one-shot manner; thus there is no need to retrain
with all the past data unlike ELM.

%
%
In this approach, the major bottleneck is the pseudo inverse operation 
${(\bm{I} + \bm{H_{i}}\bm{P_{i-1}}\bm{H_{i}^T})}^{-1}$. 
As in \cite{Tsukada18,Tsukada20}, the batch size $k$ is fixed at one
in this paper so that the pseudo inverse operation of $k \times k$
matrix for the sequential training is replaced with a simple reciprocal
operation; thus we can eliminate the SVD or QRD computation.

\subsection{Autoencoder}\label{ssec:ae}

\begin{figure}[t]
    \centering
    \includegraphics[height=50mm]{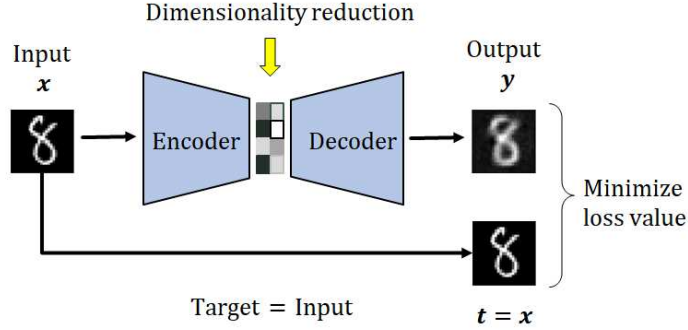}
    \caption{Autoencoder}
    \label{fig:autoencoder}
\end{figure}

%
%
Autoencoder \cite{Hinton06} is a type of neural networks developed for
dimensionality reduction, as shown in Figure \ref{fig:autoencoder}.
In this paper, OS-ELM is combined with autoencoder for unsupervised 
or semi-supervised anomaly detection.
In this case, the numbers of input- and output-layer nodes are the same
(i.e., $n=m$), while the number of hidden-layer nodes is set to less
than that of input-layer nodes (i.e., $\tilde{N}<n$). 
In autoencoder, an input chunk is converted into a well-characterized
dimensionally reduced form at the hidden layer. 
The process for the dimensionality reduction is denoted as
``encoder'', and that for decompressing the reduced form is denoted as
``decoder''. 
In OS-ELM, the encoding result for an input chunk $\bm{x}$ is obtained
as $\bm{H} = G(\bm{x} \cdot \bm{\alpha} + \bm{b})$, and the decoding
result for the hidden-layer matrix $\bm{H}$ is obtained as 
$\bm{y} = \bm{H} \cdot \bm{\beta}$.

%
%
In the training phase, an input chunk $\bm{x}$ is used as a target
chunk $\bm{t}$. 
That is, the output weight matrix $\bm{\beta}$ is trained so that an
input data is reconstructed as correctly as possible by autoencoder. 
Assuming that the model is trained with a specific input pattern,
the difference between the input data and reconstructed data (denoted
as loss value) becomes large when the input data is far from the
trained pattern.
Please note that autoencoder does not require any labeled training
data for the training phase; so it is used for unsupervised 
or semi-supervised anomaly detection.
In this case, incoming data with high loss value should be
automatically rejected before training for stable anomaly detection.


\section{On-Device Federated Learning}\label{sec:design}

%
%
As an on-device learning algorithm, in this paper, we employ a
combination of OS-ELM and autoencoder for online sequential training
and semi-supervised anomaly detection \cite{Tsukada20}.
It is further optimized by setting the batch size $k$ to one, in order
to eliminate the pseudo inverse operation of $k \times k$ matrix
for the sequential training.
A low-cost forgetting mechanism that does not require the pseudo
inverse operation is also proposed in \cite{Tsukada20}.

%
%
In practice, anomaly patterns should be accurately detected from
multiple normal patterns.
To improve the accuracy of anomaly detection in such cases, we employ
multiple on-device learning instances, each of which is specialized
for each normal pattern as proposed in \cite{Ito19}.
Also, the number of the on-device learning instances can be dynamically tuned
at runtime as proposed in \cite{Ito19}.

%
%
In this paper, the on-device learning algorithm is extended for the
on-device federated learning by applying the E$^2$LM approach to the
OS-ELM based sequential training.
In this case, edge devices can share their intermediate trained results
and update their model using those collected from the other edge
devices.
In this section, OS-ELM algorithm is analyzed so that the E$^2$LM 
approach is applied to OS-ELM for enabling the cooperative model update.
The proposed on-device federated learning approach is then
illustrated in detail.

\subsection{Modifications for OS-ELM}\label{ssec:customize_oselm}

%
%
Here, we assume that edge devices exchange the intermediate results of
their output weight matrix $\bm{\beta}$ (see Equation \ref{equation_pinv3}).
These intermediate results are obtained by $\bm{U} = \bm{H^T}\bm{H}$
and $\bm{V} = \bm{H^T}\bm{t}$, based on E$^2$LM algorithm.
Please note that the original E$^2$LM approach is designed for ELM,
which assumes a batch training, not a sequential training.
That is, $\bm{U}$ and $\bm{V}$ are computed by using the whole
training dataset.
On the other hand, our on-device learning algorithm relies on the OS-ELM
based sequential training, in which the weight matrix is
sequentially updated every time a new data comes.
If the original E$^2$LM approach is directly applied to our on-device
learning algorithm, all the past dataset must be preserved in edge
devices, which would be infeasible for resource-limited edge devices.

%
%
To address this issue, OS-ELM is analyzed as follows.
In Equation \ref{eq:os_elm_train1}, $\bm{K_i}$ is defined as 
\begin{equation}
\bm{K_i} \equiv {\begin{bmatrix} \bm{H_0} \\ \vdots \\ \bm{H_i}\end{bmatrix}}^{T}\begin{bmatrix} \bm{H_0} \\ \vdots \\ \bm{H_i}\end{bmatrix} (i \ge 0),
\end{equation} 
which indicates that it accumulates all the hidden-layer matrixes
that have been computed with up to the $\bm{i}$-th training chunk.
In this case, $\bm{U}$ and $\bm{V}$ of E$^2$LM can be computed based on
$\bm{K_i}$ and its inverse matrix $\bm{P_i}$ of OS-ELM as follows.
\begin{align}\label{equation_seq_uv}
    \bm{U_i} &= \bm{K_i} = \bm{P_i}^{-1} \nonumber \\
    \bm{V_i} &= \bm{U_i}\bm{\beta_i},
\end{align}
where $\bm{U_i}$ and $\bm{V_i}$ are intermediate results for the
$\bm{i}$-th training chunk. 
$\bm{U_i}$ and $\bm{V_i}$ can be sequentially computed with the
previous training results.
To support the on-device federated learning, Equation
\ref{equation_seq_uv} is newly inserted to the sequential training 
algorithm of OS-ELM, which will be introduced in Section
\ref{ssec:model_update}.

\subsection{Cooperative Mode Update Algorithm}\label{ssec:model_update}

\begin{figure}[t]
    \centering
    \includegraphics[height=70mm]{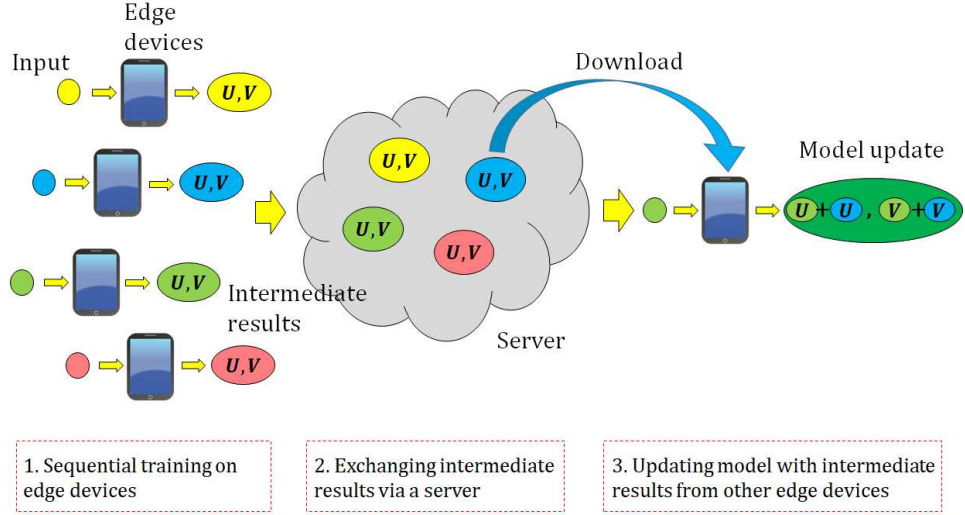}
    \caption{Overview of cooperative model update}
    \label{fig:feder}
\end{figure}

%
%
Figure \ref{fig:feder} illustrates a cooperative model update of the
proposed on-device federated learning.
It consists of the following three phases: 
\begin{enumerate}
\item Sequential training on edge devices,
\item Exchanging their intermediate results via a server, and 
\item Updating their model with necessary intermediate results
  from the other edge devices.
\end{enumerate}

%
%
First, edge devices independently execute a sequential training by
using OS-ELM algorithm.
They also compute the intermediate results $\bm{U}$ and $\bm{V}$ by
Equation \ref{equation_seq_uv}.
Second, they upload their intermediate results to a server.
We assume that the input weight matrix $\bm{\alpha}$ and the bias vector
$\bm{b}$ are the same in the edge devices.
They download necessary intermediate results from the server
if needed.
They update their model based on their own intermediate results and
those downloaded from the server by Equation \ref{equation_uv2}. 

\begin{figure}[t]
    \centering
    \includegraphics[height=70mm]{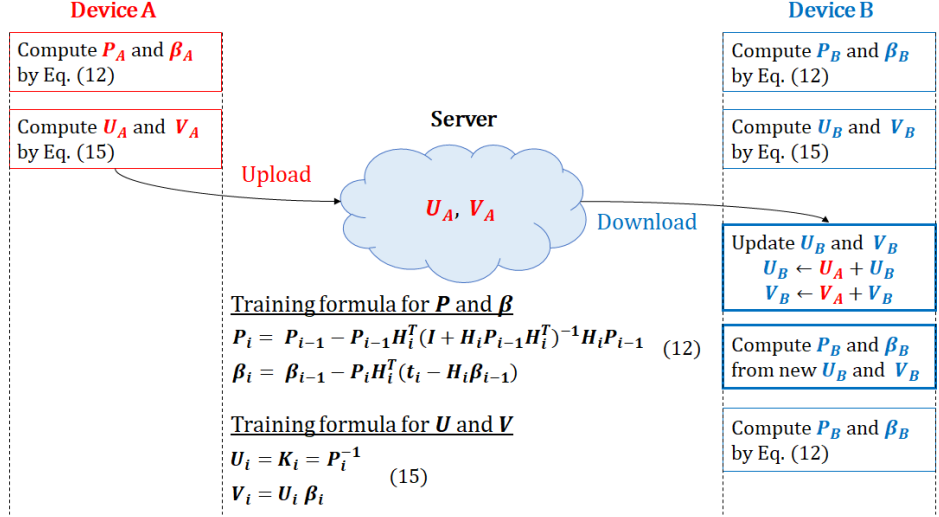}
    \caption{Flowchart of cooperative model update}
    \label{fig:feder_flow}
\end{figure}

%
%
Figure \ref {fig:feder_flow} shows a flowchart of the proposed
cooperative model update between two devices: Device-A and Device-B.
Assuming Device-A sends its intermediate results and Device-B receives
them for updating its model, their cooperative model update is
performed by the following steps.

\begin{enumerate}
\item Device-A and Device-B sequentially train their own model by
  using OS-ELM algorithm.
  In other words, they compute the output weight matrix $\bm{\beta}$
  and its intermediate result $\bm{P}$ by Equation \ref{eq:os_elm_train2}.
\item Device-A computes the intermediate results $\bm{U_A}$ and
  $\bm{V_A}$ by Equation \ref{equation_seq_uv} to share them with
  other edge devices.
  Device-B also computes $\bm{U_B}$ and $\bm{V_B}$.
  They upload these results to a server.
\item Assuming Device-B demands the Device-A's trained results, it
  downloads $\bm{U_A}$ and $\bm{V_A}$ from the server.
\item Device-B integrates their intermediate results by computing
  $\bm{U_B} \leftarrow \bm{U_A} + \bm{U_B}$ and 
  $\bm{V_B} \leftarrow \bm{V_A} + \bm{V_B}$ by using E$^2$LM algorithm.
\item Device-B updates $\bm{P_B}$ and $\bm{\beta_B}$ by computing
  $\bm{P_B} \leftarrow \bm{U_B}^{-1}$ and
  $\bm{\beta_B} \leftarrow \bm{U_B}^{-1}\bm{V_B}$.
\item Device-B can execute prediction and/or sequential training of
  OS-ELM algorithm by using the integrated $\bm{P_B}$ and $\bm{\beta_B}$.
\end{enumerate}

%
%
Edge devices can share their trained results by exchanging their
intermediate results $\bm{U}$ and $\bm{V}$ in the proposed on-device
federated learning approach, which can mitigate the privacy issues
since they do not share raw data for the cooperative model update.
Please note that the intermediate results $\bm{U}$ and $\bm{V}$ in
Equation \ref{equation_seq_uv} should be updated only when they are
sent to a server or the other edge devices; so there is no need
to update them for every input chunk. 

Regarding the client selection strategy that determines
which models of client devices are merged, in this paper we assume a
simple case where predefined edge devices share their intermediate
trained results for simplicity. 
Such client selection strategies have been studied recently. For
example, a client selection strategy that takes into account
computation and communication resource constraints is proposed for
heterogeneous edge devices in \cite{Nishio19}. 
A client selection strategy that can improve anomaly detection
accuracy by excluding unsatisfying local models is proposed in
\cite{Qin20}. 
Our proposed on-device federated learning can be combined with these
client selection strategies in order to improve the accuracy or
efficiency, though such a direction is our future work.


\section{Evaluations}\label{sec:eval}

First, the behavior of the proposed on-device federated learning
approach is demonstrated by merging trained results from multiple edge
devices in Section \ref{ssec:eval_result}.
Then, prediction results using the merged model are compared to
those produced by traditional 3-layer BP-NN (backpropagation based
neural network) and 5-layer BP-NN in terms of the loss values and
ROC-AUC (Receiver Operating Characteristic Curve - Area Under Curve)
scores in Section \ref{ssec:eval_auc}.
 Those are also compared to a traditional BP-NN based
federated learning approach. 
In addition, the proposed on-device federated learning is evaluated in
terms of the model merging latency in Section \ref{ssec:latency}, and
it is compared to a conventional sequential training approach in 
Section \ref{ssec:converge}.
Table \ref{tb:env} shows specification of the experimental machine.

\begin{table}[t]
    \centering
    \caption{Specification of experimental machine}
    \label{tb:env}
    \begin{tabular}{l|c} \hline \hline
            OS & Ubuntu 17.10 \\
            CPU & Intel Core i5 7500 3.4GHz \\
            DRAM & 8GB \\
            Storage & SSD 480GB \\\hline    
    \end{tabular}
\end{table}

\begin{table}[t]
    \centering
    \caption{Three datasets}
    \label{tb:dataset}
    \begin{tabular}{l|c|c} \hline \hline
            Name & Features & Classes \\\hline
            UAH-DriveSet \cite{uah_dataset} & 225 & 3 \\
            Smartphone HAR \cite{har} & 561 & 6 \\
            MNIST \cite{mnist} & 784 & 10 \\\hline  
    \end{tabular}
\end{table}

\subsection{Evaluation Environment}\label{ssec:eval_env}

\subsubsection{Datasets}
The evaluations are conducted with three datasets shown in Table
\ref{tb:dataset}.
 Throughout this evaluation, we assume a semi-supervised
anomaly detection approach that constructs a model from normal
patterns only.
In other words, the trained patterns are regarded as normal and the
others are anomalous.
In this case, a loss value should be low for normal data while it
should be high for anomalous data. 

%
%
UAH-DriveSet \cite{uah_dataset} contains car driving histories of six
drivers simulating three different driving patterns: aggressive,
drowsy, and normal.
It can be used for the aggressive driving detection.
Their car speed was extracted from the GPS data obtained from a
smartphone fixed in their cars.
The sampling frequency of the car speed was 1Hz.
In our experiment, the car speed is quantized with 15 levels
(1 level = 10km/h).
Assuming a state is assigned to each speed level, we can build a
state-transition probability table with 15$\times$15 entries, each of
which represents a probability of a state transition from one state to
another.

%
%
Smartphone HAR dataset \cite{har} contains human activity recordings
of 30 volunteers simulating six activities: walking, walking\_upstairs,
walking\_downstairs, sitting, standing, and laying.
It can be used for human activity recognition.
Their 3-axial linear acceleration and 3-axial angular velocity were
captured at a constant rate of 50Hz using waist-mounted smartphone
with embedded inertial sensors.
The sensor signals were pre-processed.
As a result, 561 features consisting of time and frequency domain
variables were obtained for each time-window.

%
%
MNIST dataset \cite{mnist} contains handwritten digits from 0 to 9.
It is widely used for training and testing 
in various fields of machine learning.
Each digit size is 28$\times$28 pixel in gray scale, 
resulting in 784 features.
In our experiment, 
all the pixel values are divided by 255 
so that they are normalized to [0, 1].

\subsubsection{Setup}
%
%
A vector of 225 features from the car driving dataset, that of 561
features from the human activity dataset, and that of 784 features 
from MNIST dataset are fed to
the neural-network based on-device learning algorithm 
\cite{Tsukada20} for anomaly detection.
The numbers of input-layer nodes 
and output-layer nodes are same in all the experiments. 
The forget factor $\alpha$ is 1 (i.e., no forgetting).
The batch size $k$ is fixed to 1.
The number of training epochs $E$ is 1.
The number of anomaly detection instances is 2 \cite{Ito19}.
These instances are denoted as Device-A and Device-B in our experiments.
The OS-ELM based anomaly detection with the proposed cooperative model
update is implemented with Python 3.6.4 and NumPy 1.14.1.
%
%
As a comparison with the proposed OS-ELM based anomaly detection, 
a 3-layer BP-NN based autoencoder 
and 5-layer BP-NN based deep autoencoder 
(denoted as BP-NN3 and BP-NN5, respectively) are implemented 
with TensorFlow v1.12.0 \cite{tensorflow}.
The hyperparameter settings in OS-ELM, BP-NN3, and BP-NN5 are
listed in Table \ref{tb:eval_hp}\footnote{
    $G_{hidden}$: activation function applied to all the hidden layers.
    $G_{out}$: activation function applied to the output layer.
    $p(x)$: probability density function used for random initialization 
of input weight $\bm{\alpha}$ and bias vector $\bm{b}$ in OS-ELM.
    $\tilde{N}_{i}$: the number of nodes of the $i$th hidden layer.
    $L$: loss function.
    $O$: optimization algorithm.
    $k$: batch size.
    $E$: the number of training epochs.
}.
Here, 10-fold cross-validation for ROC-AUC criterion 
is conducted to tune the hyperparameters with each dataset.

\begin{table}[t]
    \centering
    \caption{Hyperparameter settings}
    \label{tb:eval_hp}
    \begin{tabular}{c|c}\hline\hline
    \multicolumn{2}{c}{OS-ELM: $\{G_{hidden}, p(x), \tilde{N}_1, L\}$}  \\\hline
    UAH-DriveSet & $\{$Sigmoid, Uniform, 16, MSE\footnotemark$\}$  \\ 
    Smartphone HAR & $\{$Identity\footnotemark, Uniform, 128, MSE$\}$  \\
    MNIST & $\{$Identity, Uniform, 64, MSE$\}$ \\\hline
    \multicolumn{2}{c}{BP-NN3: $\{G_{hidden}, G_{out}, \tilde{N}_1, L, O, k, E\}$} \\\hline
    Smartphone HAR & $\{$Relu, Sigmoid, 256, MSE, Adam, 8, 20$\}$  \\
    MNIST & $\{$Relu, Sigmoid, 64, MSE, Adam, 32, 5$\}$  \\\hline
    \multicolumn{2}{c}{BP-NN5: $\{G_{hidden}, G_{out}, \tilde{N}_1, \tilde{N}_2, \tilde{N}_3, L, O, k, E\}$} \\\hline
    Smartphone HAR &  $\{$Relu, Sigmoid, 128, 256, 128, MSE, Adam, 8, 20$\}$ \\
    MNIST &  $\{$Relu, Sigmoid, 64, 32, 64, MSE, Adam, 8, 10$\}$ \\\hline
    \end{tabular}
\end{table}
\footnotetext[3]{$L(\bm{x}, \bm{y}) = \frac{1}{n}\sum_{i=0}^{n}(x_i - y_i)^2$}
\footnotetext[4]{$G(\bm{x}) = \bm{x}$}

\subsection{Loss Values Before and After Model Update}\label{ssec:eval_result}
\subsubsection{Setup}
%
%
Here, we compare loss values before and after the cooperative model
update.
Below is the experiment scenario using the two instances with the car
driving dataset.
\begin{enumerate}
\item Device-A trains its model so that the normal driving pattern
  becomes normal,  and the others are anomalous. 
  Device-B trains its model so that the aggressive driving pattern
  becomes normal,  and the others are anomalous.
\item Aggressive and normal driving patterns are fed to Device-A to
  evaluate the loss values.
\item Device-B uploads its intermediate results to a server, and
  Device-A downloads them from the server.
\item Device-A updates its model based on its own intermediate results
  and those from Device-B.
   It is expected that Device-B's normal becomes normal at
  Device-A. 
\item The same testing as Step 2 is executed again.
\end{enumerate}
%
%
Below is the experiment scenario using the human activity dataset.
\begin{enumerate}
\item Device-A trains its model so that the sitting pattern becomes
  normal,  and the others are anomalous. 
  Device-B trains its model so that the laying pattern becomes
  normal,  and the others are anomalous. 
\item Walking, walking\_upstairs, walking\_downstairs, sitting,
  standing, and laying patterns are fed to Device-B to evaluate the
  loss values.
\item Device-A uploads its intermediate results to a server, and
  Device-B downloads them from the server.
\item Device-B updates its model based on its own intermediate results
  and those from Device-A.
   It is expected that Device-A's normal becomes normal at
  Device-B. 
\item The same testing as Step 2 is executed again.
\end{enumerate}

\begin{figure}[t]
    \centering
    \includegraphics[height=65mm]{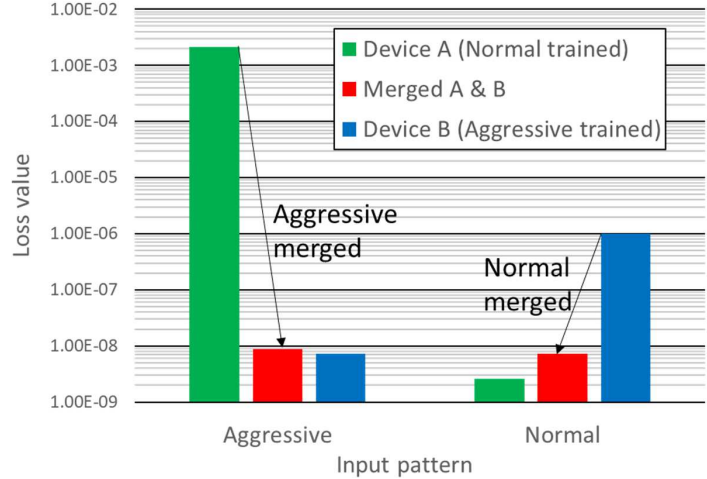}
    \caption{Loss values before and after cooperative model update with car driving dataset}
    \label{fig:feder_eval}
\end{figure}

\begin{figure*}[t]
    \centering
    \includegraphics[height=75mm]{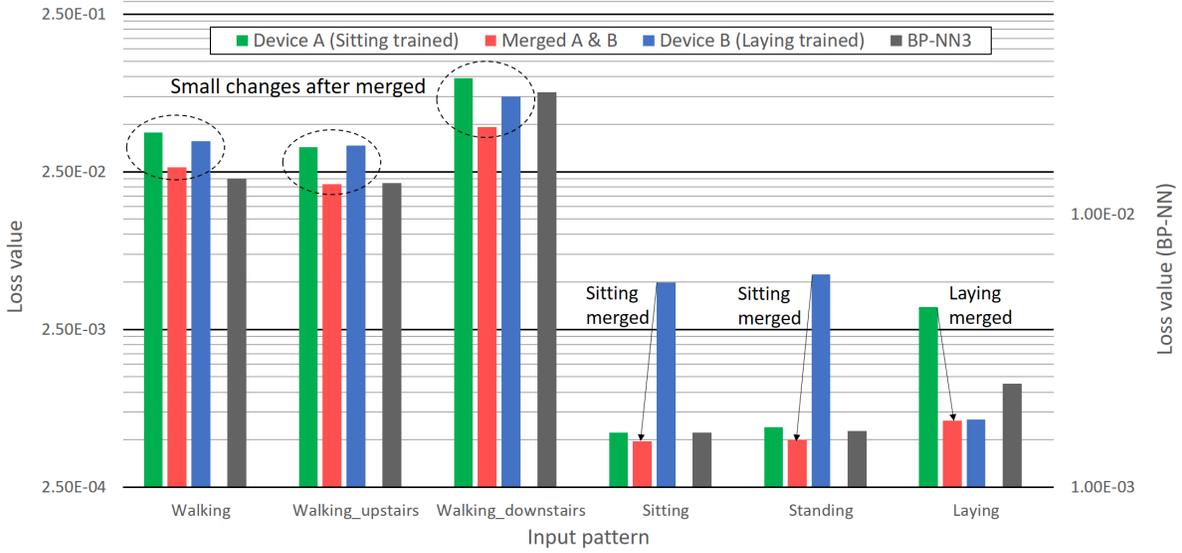}
    \caption{Loss values before and after cooperative model update with human activity dataset}
    \label{fig:har_eval}
\end{figure*}

%
%
In these scenarios, the loss values at Step 2 are denoted as
``before the cooperative model update''.
Those at Step 5 are denoted as ``after the cooperative model
update''.
In this setup, after the cooperative model update, ``Device-A that
has merged Device-B'' and ``Device-B that has merged Device-A'' are
identical.
A low loss value means that a given input pattern is well
reconstructed by autoencoder, which means that the input pattern is
normal in the edge device.
In the first scenario, Device-A is adapted to the aggressive and normal
driving patterns with the car driving dataset.
In the second one, Device-B is adapted to the sitting and laying patterns
with the human activity dataset.

\subsubsection{Results}
%
%
Figure \ref{fig:feder_eval} shows the loss values before and after the
cooperative model update with the car driving dataset.
X-axis represents the input patterns.
Y-axis represents the loss values in a logarithmic scale.
Green bars represent loss values of Device-A before the cooperative
model update, while red bars represent those after the cooperative
model update.
Blue bars represent loss values of Device-B.
In the case of aggressive pattern, the loss value of Device-A before
the cooperative model update (green bar) is high, because Device-A is
trained with the normal pattern.
The loss value then becomes quite low after integrating the intermediate
results of Device-B to Device-A (red bar).
This means that the trained result of Device-B is correctly added
to Device-A.
In the case of normal pattern, the loss value before merging (green
bar) is low, but it slightly increases after the trained result
of Device-B is merged (red bar).
Nevertheless, the loss value is still quite low.
We can observe the same tendency for Device-B by comparing the blue
and red bars.

%
%
Figure \ref{fig:har_eval} shows the loss values before and after 
the cooperative model update with the human activity dataset.
Regarding the loss values, the same tendency with the driving dataset
is observed.
In the case of sitting pattern, the loss value of Device-B before
the cooperative model update (blue bar) is high, because Device-B is
trained with the laying pattern.
Then, the loss value becomes low after the trained result of 
Device-A is merged (red bar). 
In the case of laying pattern, the loss value of Device-A before
merging (green bar) is high and significantly decreased after merging
of the trained result of Device-B (red bar).
On the other hand, in the walking, walking\_upstairs, and
walking\_downstairs patterns, their loss values before and after the
cooperative model update are relatively close.
These input patterns are detected as anomalous even after the
cooperative model update, because they are not normal for both
Device-A and Device-B.   
In the case of standing pattern, the similar tendency as the sitting
pattern is observed.
The loss value becomes low after the trained result of Device-A is
merged to Device-B.
This means that there is a similarity between the sitting pattern and
standing pattern.

%
%
As a counterpart of the proposed OS-ELM based anomaly detection, 
a 3-layer BP-NN based autoencoder is implemented (denoted as BP-NN3).
BP-NN3 is trained with the sitting pattern and
laying pattern.
In Figure \ref{fig:har_eval}, gray bars (Y-axis on the right side)
represent loss values of BP-NN3 in a logarithmic
scale.
Please note that absolute values of its loss values are different from
OS-ELM based ones since their training algorithms are different.
Nevertheless, the tendency of BP-NN3 (gray bars)
is very similar to that of the proposed cooperative model update
(red bars).
 This means that Device-B's model after the trained result of 
Device-A is merged can distinguish between normal and anomalous
input patterns as accurately as the BP-NN based autoencoder. 

\begin{figure}[t]
    \centering
    \includegraphics[height=50mm]{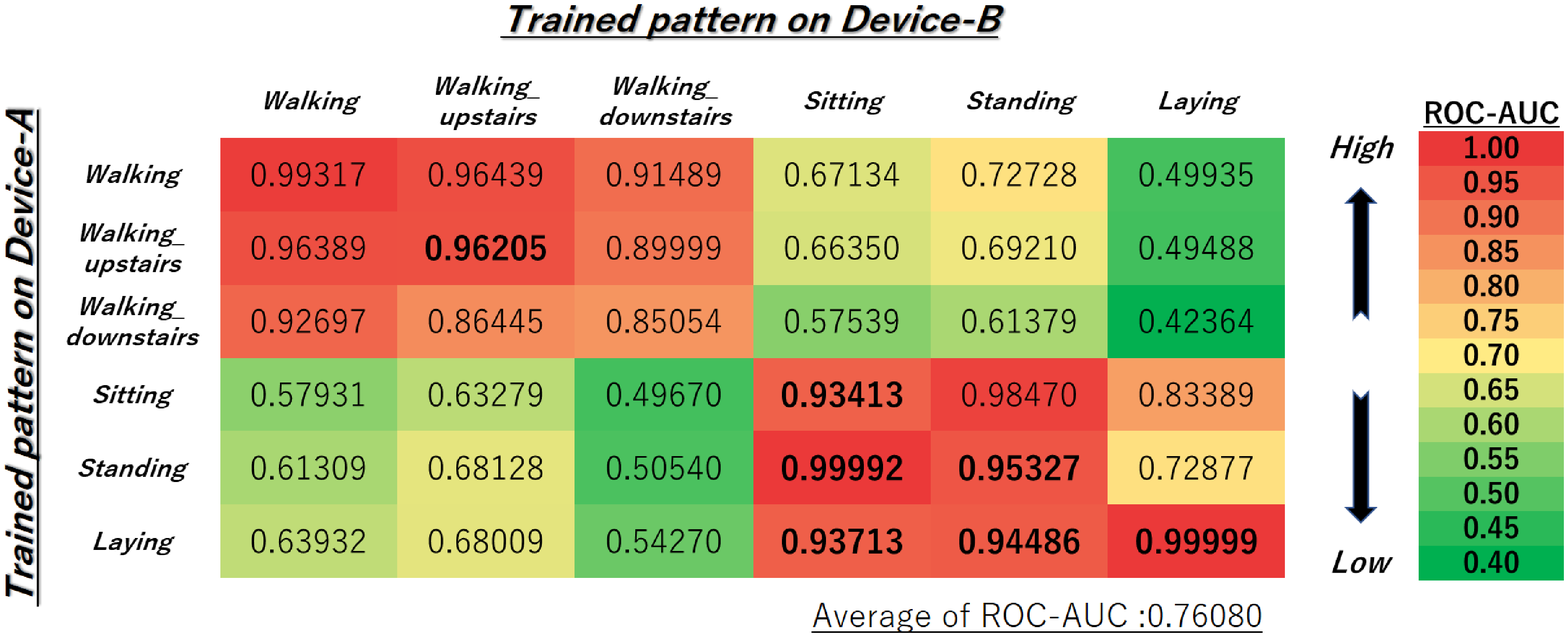}
    \caption{ROC-AUC scores before cooperative model update with human activity dataset}
    \label{fig:har_nonFL}
    \centering
    \includegraphics[height=50mm]{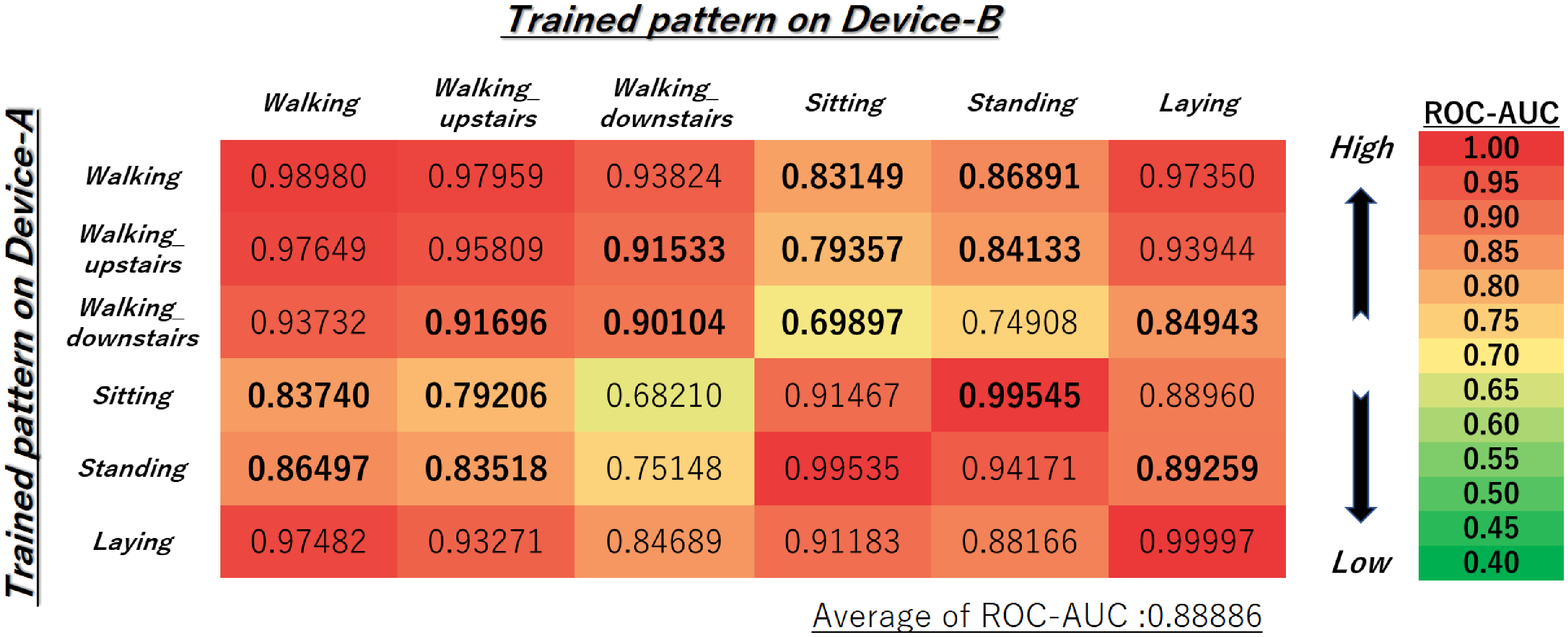}
    \caption{ROC-AUC scores after cooperative model update with human activity dataset}
    \label{fig:har_FL}
    \centering
    \includegraphics[height=50mm]{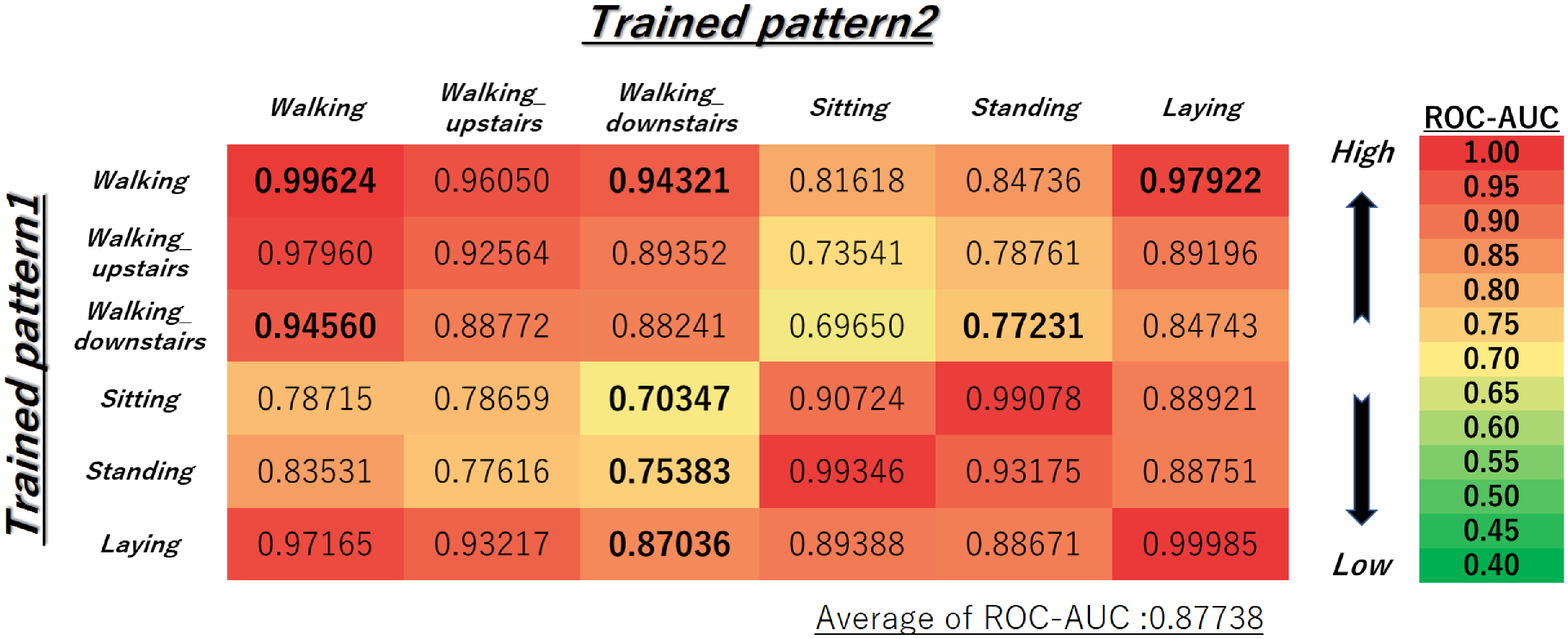}
    \caption{ROC-AUC scores of BP-NN3 with human activity dataset}
    \label{fig:har_BP3}
\end{figure}
\begin{figure}[t]
    \centering
    \includegraphics[height=50mm]{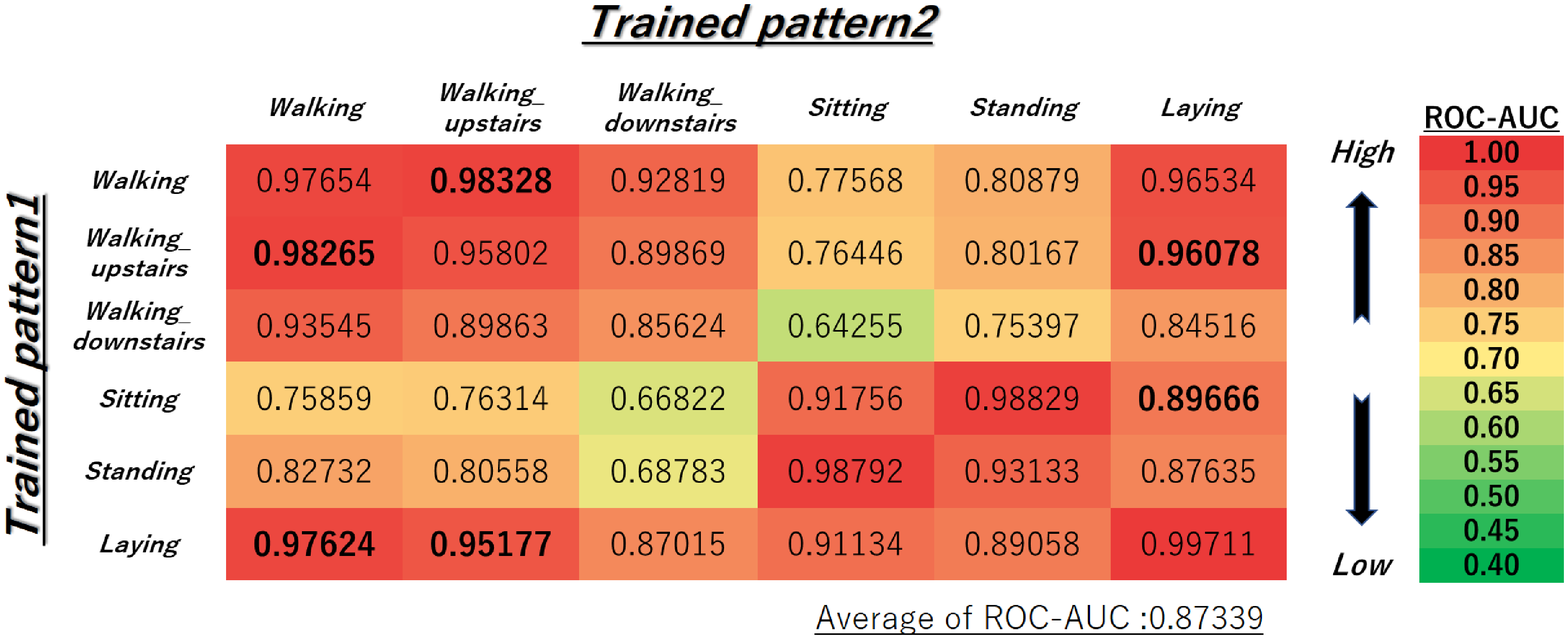}
    \caption{ROC-AUC scores of BP-NN5 with human activity dataset}
    \label{fig:har_BP5}
    \centering
    \includegraphics[height=50mm]{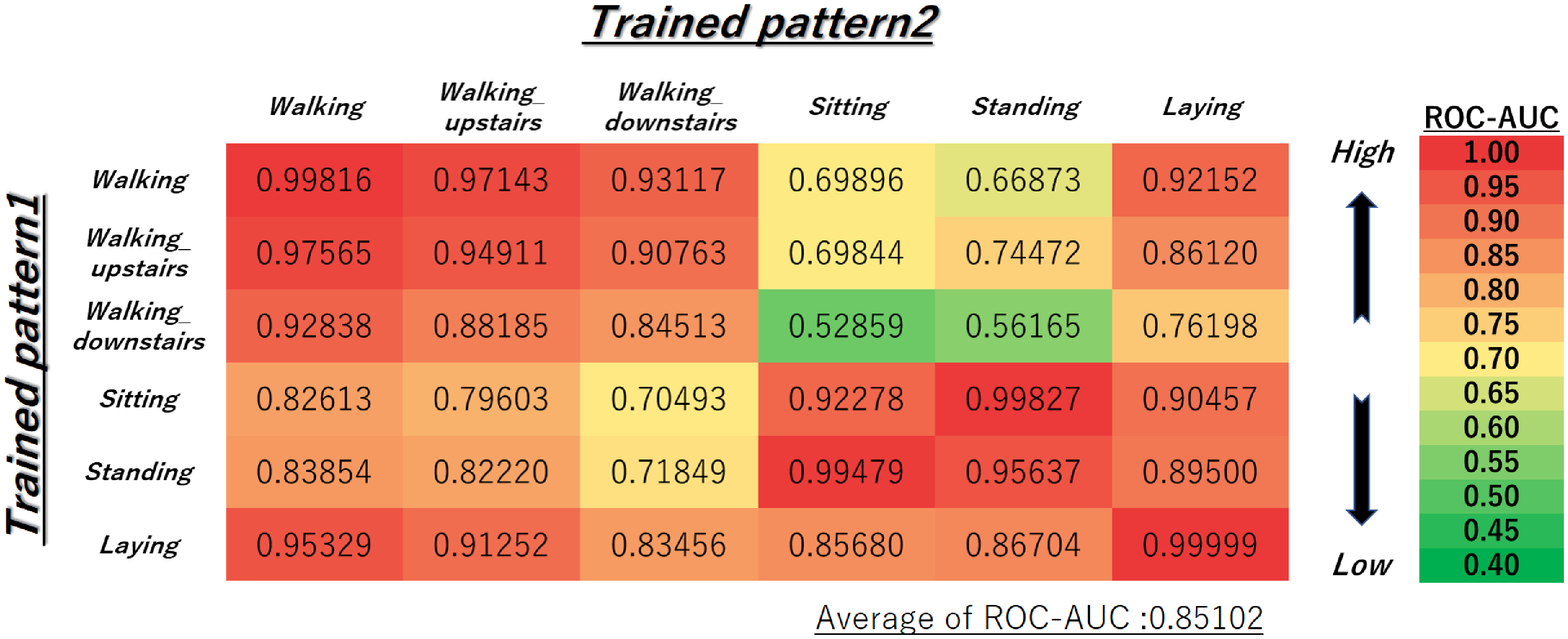}
    \caption{ROC-AUC scores of traditional FL (BP-NN3-FL) with human activity dataset}
    \label{fig:har_traditionalFL}
\end{figure}

\subsection{ROC-AUC Scores Before and After Model Update}\label{ssec:eval_auc}
\subsubsection{Setup}
%
%
Here, ROC-AUC scores 
before and after the cooperative model update are compared 
using the human activity dataset and MNIST dataset.
The following five steps are performed 
for every combination of two patterns 
(denoted as $p_A$ and $p_B$) in each dataset.
\begin{enumerate}
\item Device-A trains its model so that $p_A$ becomes normal,
   and the others are anomalous. 
  Device-B trains its model so that $p_B$ becomes normal,
   and the others are anomalous. 
\item ROC-AUC scores are evaluated using all the patterns on Device-A.
\item Device-B uploads its intermediate results to a server, and
  Device-A downloads them from the server.
\item Device-A updates its model based on its own intermediate results
  and those from Device-B.
   It is expected that Device-B's normal becomes normal at
  Device-A. 
\item The same testing as Step 2 is executed again.
\end{enumerate}

\begin{figure*}[t]
    \centering
    \includegraphics[height=50mm]{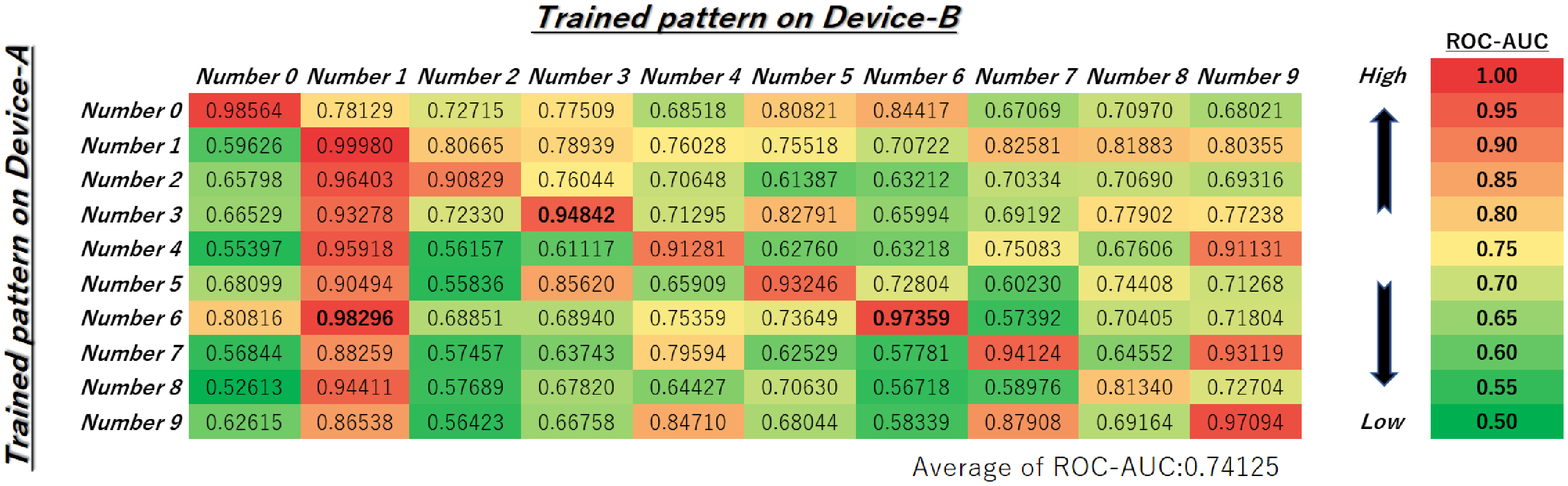}
    \caption{ROC-AUC scores before cooperative model update with MNIST dataset}
    \label{fig:mnist_nonFL}
    \centering
    \includegraphics[height=50mm]{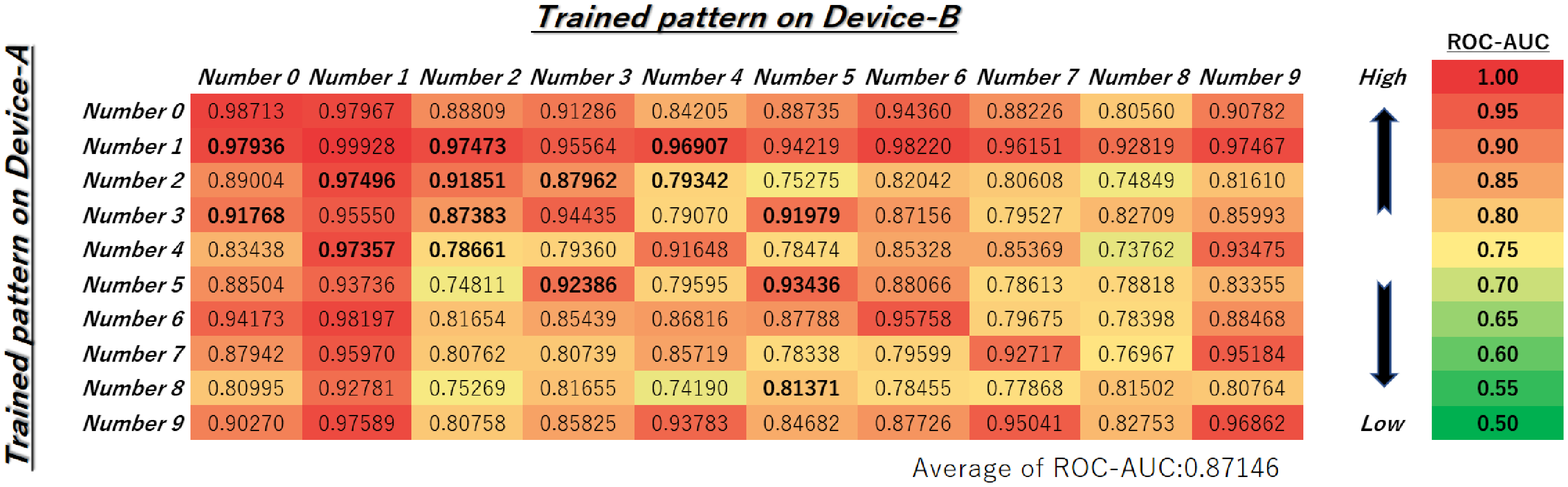}
    \caption{ROC-AUC scores after cooperative model update with MNIST dataset}
    \label{fig:mnist_FL}
    \centering
    \includegraphics[height=50mm]{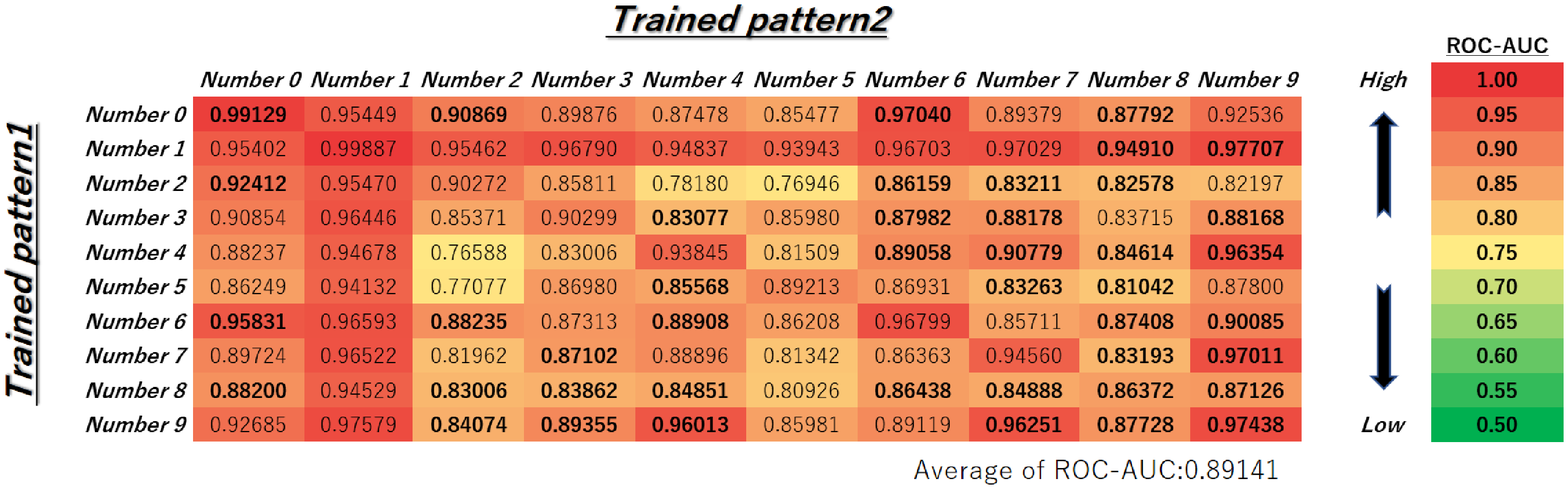}
    \caption{ROC-AUC scores of BP-NN3 with MNIST dataset}
    \label{fig:mnist_BP3}
\end{figure*}
\begin{figure*}[t]
    \centering
    \includegraphics[height=50mm]{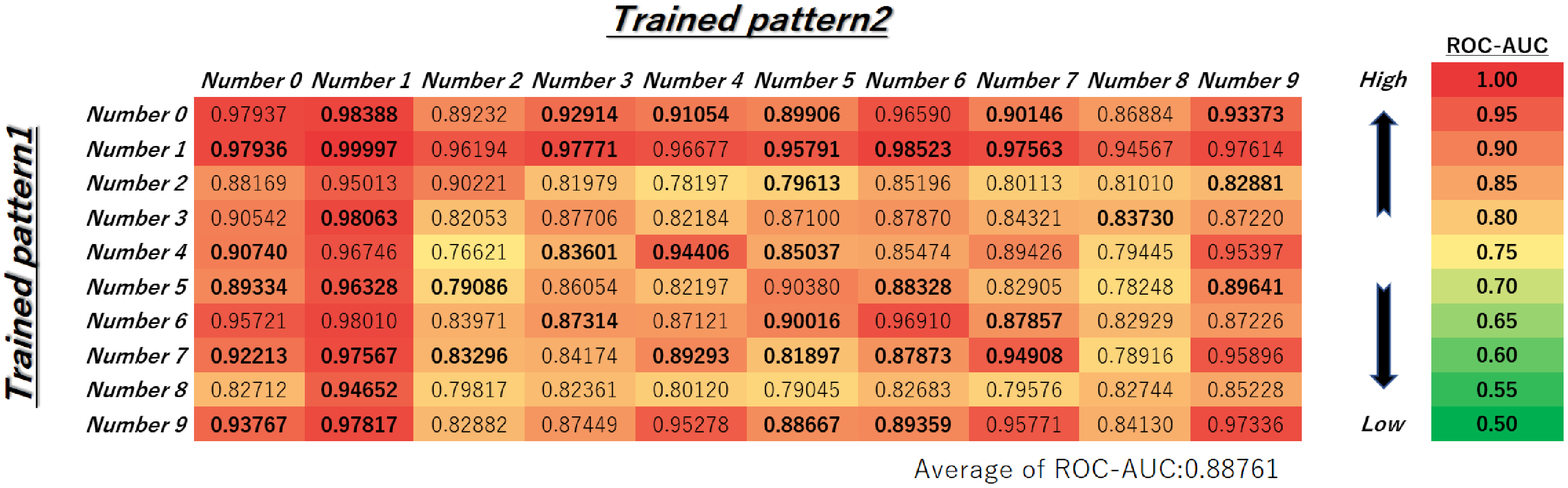}
    \caption{ROC-AUC scores of BP-NN5 with MNIST dataset}
    \label{fig:mnist_BP5}
    \centering
    \includegraphics[height=50mm]{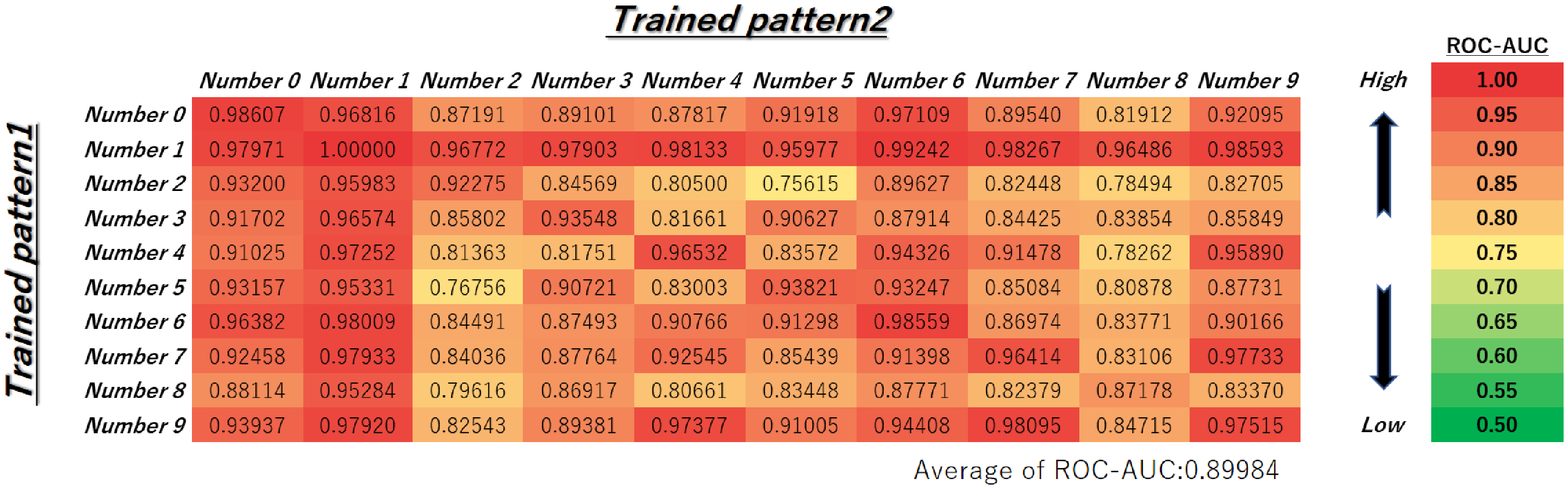}
    \caption{ROC-AUC scores of traditional FL (BP-NN3-FL) with MNIST dataset}
    \label{fig:mnist_traditionalFL}
\end{figure*}

Trained patterns $p_A$ and $p_B$ in Step 1 are used as normal test data
in Step 2 and Step 5.
Whereas, the other untrained patterns are used as anomalous test data.
For example, using the human activity dataset, 
when Device-A is trained with the walking pattern
and Device-B is trained with the standing pattern in Step 1, 
the walking and standing patterns are used as normal data; 
on the other hand, the other patterns 
(i.e., walking\_upstairs, walking\_downstairs, sitting, and laying patterns) 
are used as anomalous data in Step 2 and Step 5.
ROC-AUC scores are evaluated for every combination of patterns 
with the human activity dataset and MNIST dataset.
ROC-AUC scores at Step 2 are denoted as
``before the cooperative model update''.
Those at Step 5 are denoted as 
``after the cooperative model update''.

%
%
The proposed OS-ELM based federated learning approach is compared to
a 3-layer BP-NN based autoencoder (BP-NN3) and a 5-layer BP-NN based
deep autoencoder (BP-NN5).
BP-NN3 and BP-NN5 train their model so that every combination of two
patterns becomes normal.
In the case of BP-NN based autoencoders, the two trained patterns are
used as normal test data, while the others are used as anomalous test
data to evaluate ROC-AUC scores. 
ROC-AUC scores are calculated for every combination of two patterns in
each dataset.
 In addition, a traditional federated learning approach
using BP-NN3 (denoted as BP-NN3-FL) is implemented.
In each communication round, two patterns are trained separately based on 
a single global model.
Then, these locally trained models are averaged, and the global model
is updated, which will be used for local train of the next round.
The number of communication rounds $R$ is set to
50 in all the datasets for stable anomaly detection performance in BP-NN3-FL.
Note that $R$ versus accuracy is well analyzed in \cite{Brendan16}.
Its ROC-AUC scores are calculated as well as BP-NN3 and BP-NN5. 

%
%
ROC-AUC is widely used as a criterion for 
evaluating the model performance of anomaly detection
independently of particular anomaly score thresholds.
ROC-AUC scores range from 0 to 1.
A higher ROC-AUC score means that 
the model can detect both the normal and anomalous patterns 
more accurately.
In this experiment, 80\% of samples are used as training data 
and the others are used as test data in each dataset.
The number of anomaly samples in the test dataset is limited to 
10\% of that of normal samples.
The final ROC-AUC scores are averaged over 50 trials
for every combination of patterns in each dataset.

\subsubsection{Results}
%
%
Figures \ref{fig:har_nonFL}-\ref{fig:har_traditionalFL} show ROC-AUC
scores with the human activity dataset using heat maps.
 The highest score among the five models 
(before and after the cooperative model update, BP-NN3, BP-NN5, and 
BP-NN3-FL) is shown in bold. 
Figures \ref{fig:har_nonFL} and \ref{fig:har_FL} show the results of 
before and after the proposed cooperative model update. 
In these graphs, each row represents a trained pattern on Device-A, 
while each column represents a trained pattern on Device-B.
Figures \ref{fig:har_BP3}-\ref{fig:har_traditionalFL} show the results of 
the BP-NN based models, 
and two trained patterns are corresponding to the row and column.
In Figure \ref{fig:har_nonFL}, ROC-AUC scores 
before the proposed cooperative model update are low 
especially when trained patterns 
on Device-A and Device-B have mutually distant features.
This is because Device-A is trained with one activity pattern
and unseen patterns 
during the training phase should be detected as anomalous.
The ROC-AUC scores are then significantly increased overall 
after integrating the intermediate results of Device-B to Device-A, 
as shown in Figure \ref{fig:har_FL}.
This means that the trained result of Device-B 
is correctly added to Device-A
so that Device-A can extend the coverage of normal patterns  
in all the combinations of patterns.

%
%
In the cases of BP-NN based models shown in Figures 
\ref{fig:har_BP3}-\ref{fig:har_traditionalFL},
their tendencies and overall averages of ROC-AUC scores are
very similar to those after the proposed cooperative model update.
This means that the proposed cooperative model update can produce a
merged model by integrating trained results from the other edge
devices as accurately as  BP-NN3, BP-NN5, and BP-NN3-FL 
 in terms of ROC-AUC criterion.
Please note that these BP-NN based models need to be iteratively
trained for some epochs in order to obtain their best generalization
performance, e.g., they were trained for 20 epochs in BP-NN3 and
BP-NN5.
In contrast, the proposed OS-ELM based federated learning approach 
can always compute the optimal output weight matrix only in a single
epoch.

%
%
Figures \ref{fig:mnist_nonFL}-\ref{fig:mnist_traditionalFL} show
ROC-AUC scores with MNIST dataset.
We can observe the same tendency
with the human activity dataset
in the four anomaly detection models.
In Figure \ref{fig:mnist_nonFL}, ROC-AUC scores 
before the proposed cooperative mode update 
are low overall except for the diagonal elements, 
because Device-A is trained with one handwritten digit so that  
the others should be detected as anomalous on Device-A.
Then, the ROC-AUC scores become high even in elements other than
the diagonal ones after the trained results of Device-B are merged,
as shown in Figure \ref{fig:mnist_FL}.
 Moreover, a similar tendency as ROC-AUC scores after the
proposed cooperative model update is observed in BP-NN3, BP-NN5, and
BP-NN3-FL, 
though average ROC-AUC scores of BP-NN3, BP-NN5, and BP-NN3-FL
are slightly higher than those of the proposed cooperative model
update, 
as shown in Figures \ref{fig:mnist_BP3}-\ref{fig:mnist_traditionalFL}.
This means that the merged model on Device-A has obtained a comparable
anomaly detection performance as the BP-NN based models
with MNIST dataset.

\subsection{Training, Prediction, and Merging Latencies}\label{ssec:latency}
\subsubsection{Setup}
In this section, the proposed on-device federated learning is
evaluated in terms of training, prediction, and merging latencies
with the human activity dataset.
 In addition, these latencies are compared with those of
the BP-NN3-FL based autoencoder.
The batch size $k$ of BP-NN3-FL is set to 1 for a fair comparison with the
proposed OS-ELM based federated learning approach. 
They are compared in terms of the following latencies.
\begin{itemize}
\item Training latency is an elapsed time from receiving an input
  sample until the parameter is trained by using OS-ELM or BP-NN3-FL.
\item Prediction latency is an elapsed time from receiving an
  input sample until its loss value is computed by using OS-ELM or BP-NN3-FL.
\item Merging latency of OS-ELM is an elapsed time from receiving
  intermediate results $\bm{U}$ and $\bm{V}$ until a model update with
  the intermediate results is finished.
   That of BP-NN3-FL includes latencies for receiving two locally
  trained models, averaging them, and optimizing a global model based
  on the result. It is required for each communication round. 
\end{itemize}
These latencies are measured on the experimental machine shown in
Table \ref{tb:env}.

\begin{table}[t]
\centering
\caption{Training, prediction, and merging latencies [msec]}
\label{tb:latency}
\begin{tabular}{l|c|c|c} \hline \hline
	\multicolumn{4}{c}{Number of hidden-layer nodes $\tilde{N}=64$} \\ \hline
	& Training latency & Prediction latency & Merging latency \\ \hline
	OS-ELM    & \textbf{0.471}   & 0.089             & \textbf{5.78} \\
	 BP-NN3-FL  & 0.588 & 0.290  &  $1.95 \times R$  \\ \hline
	\multicolumn{4}{c}{Number of hidden-layer nodes $\tilde{N}=128$} \\ \hline
	& Training latency & Prediction latency & Merging latency \\ \hline
	OS-ELM    & \textbf{0.794}   & 0.106             & \textbf{21.8} \\
	 BP-NN3-FL  & 0.980 & 0.364  &  $2.42 \times R$  \\ \hline
\end{tabular}
\end{table}

\subsubsection{Results}
%
%
Table \ref{tb:latency} shows the evaluation results in the cases of 
$\tilde{N}=64$ and $\tilde{N}=128$.
The number of input features is 561.
The merging latency of OS-ELM is higher than those of training and prediction
latencies, and it depends on the number of hidden-layer nodes
because of the inverse operations of $\tilde{N}\times\tilde{N}$  
(size of matrix $\bm{U}$ is $\tilde{N}\times\tilde{N}$).
Nevertheless, the merging latency is still modest.
Please note that the merging latency of BP-NN3-FL is required for
each communication round during a training phase, while the merging 
process of our OS-ELM based federated learning approach is executed only once
(i.e., ``one-shot'').
Thus, the proposed federated learning approach is light-weight 
in terms of computation and communication costs. 

\subsection{Convergence of Loss Values}\label{ssec:converge}
\subsubsection{Setup}
%
%
The proposed cooperative model update can merge trained results of
different input patterns at a time.
On the other hand, the original OS-ELM can intrinsically adapt to new
normal patterns by continuously executing the sequential training of
the new patterns.
These two approaches (i.e., the proposed merging and the conventional
sequential training) are evaluated in terms of convergence of loss
values for a new normal pattern using the human activity dataset.
The number of hidden-layer nodes $\tilde{N}$ is 128.

%
%
In this experiment, Device-A trains its model so that the laying
pattern becomes normal, and Device-B trains its model so that the
walking pattern becomes normal.
In the proposed merging, the trained result of Device-A is integrated
to Device-B so that the laying pattern becomes normal in Device-B.
In the case of the conventional sequential training, Device-B
continuously executes sequential training of the laying pattern, so
that the loss value of the laying pattern is gradually decreased.
Its decreasing loss value is evaluated at every 50 sequential updates
and compared to that of the proposed merging.

\begin{figure}[t]
    \centering
    \includegraphics[height=70mm]{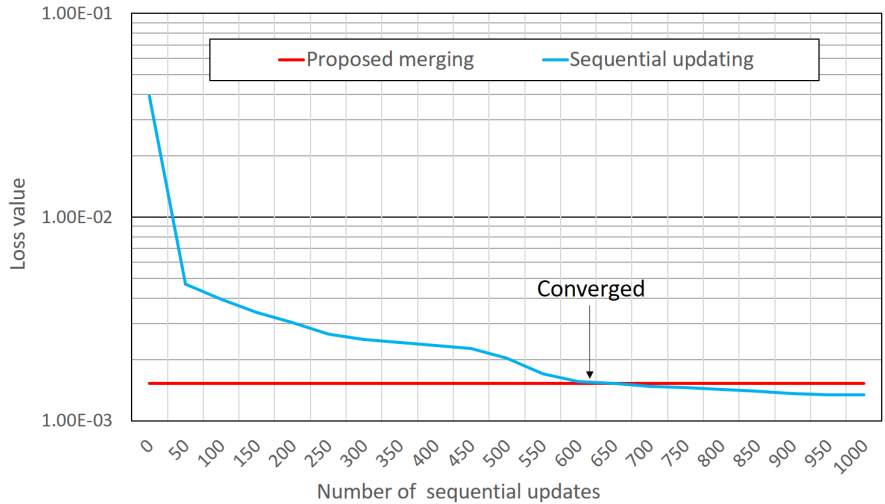}
    \caption{Convergence of loss values of merging and sequential updating}
    \label{fig:convergence}
\end{figure}

\subsubsection{Results}
%
%
Figure \ref{fig:convergence} shows the results.
X-axis represents the number of sequential updates in the conventional
sequential training.
Y-axis represents loss values of the laying pattern in a logarithmic
scale.
Red line represents the loss value of Device-B after the proposed
merging; thus, the loss value is low and constant.
Blue line represents the loss value of Device-B when sequentially
updating its model by the laying pattern; thus, the loss value is
decreased as the number of sequential updates increases.
Then, the loss value becomes as low as that of the merged one (red line) when
the number of sequential updates is approximately 650.
For 650 sequential updates, at least $0.794 \times 650$ msec is
required for the convergence, while the proposed cooperative model
update (i.e., merging) requires only 21.8 msec.
Thus, the proposed cooperative model update can merge the
trained results of the other edge devices rapidly.


\section{Conclusions}\label{sec:conc}

In this paper, we focused on a neural-network based on-device learning
approach so that edge devices can train or correct their model based
on incoming data at runtime in order to adapt to a given environment.
Since a training is done independently at distributed edge devices,
the issue is that only a limited amount of training data can be used
for each edge device.
To address this issue, in this paper, the on-device learning algorithm
was extended for the on-device federated learning by applying the
E$^2$LM approach to the OS-ELM based sequential training.
In this case, edge devices can share their intermediate trained
results and update their model using those collected from the other
edge devices.
We illustrated an algorithm for the proposed cooperative model update.
Evaluation results using the car driving dataset, the
human activity dataset, and MNIST dataset demonstrated that the
proposed on-device federated learning approach can produce a merged
model by integrating trained results from multiple edge devices as
accurately as BP-NN3, BP-NN5, and BP-NN3-FL.
Please note that the proposed approach is one-shot, which is favorable
especially in the federated learning settings since the number of
communication rounds significantly affects the
communication cost.
As a future work, we will explore client selection strategies for our
approach in order to further improve the accuracy and efficiency.


\end{document}